%% file: acl_latex.tex
\pdfoutput=1

\documentclass[11pt]{article}

\usepackage[]{acl}

\usepackage{times}
\usepackage{latexsym}
\usepackage{booktabs}
\usepackage[ruled,linesnumbered]{algorithm2e}
\usepackage{algorithmic}
\usepackage{tabularx}  
\usepackage{multirow}

\usepackage[T1]{fontenc}

\usepackage[utf8]{inputenc}

\usepackage{microtype}

\usepackage{inconsolata}

\usepackage{graphicx}
\usepackage{amssymb}
\usepackage{enumitem}
\usepackage{amsmath}
\usepackage{stfloats}
\title{LDGen: Enhancing Text-to-Image Synthesis via Large Language Model-Driven Language Representation}

\author{Pengzhi Li, Pengfei Yu$^{{\dag}}$, Zide Liu, Wei He,Xuhao Pan,
\\ 
\textbf{Xudong Rao, Tao Wei, Wei Chen$^{{\S}}$}\\
Li Auto Inc.
}

\begin{document}
\twocolumn[{
\renewcommand\twocolumn[1][]{#1}
\maketitle
\begin{center}
    \captionsetup{type=figure}
    	\includegraphics[width=0.95\linewidth]{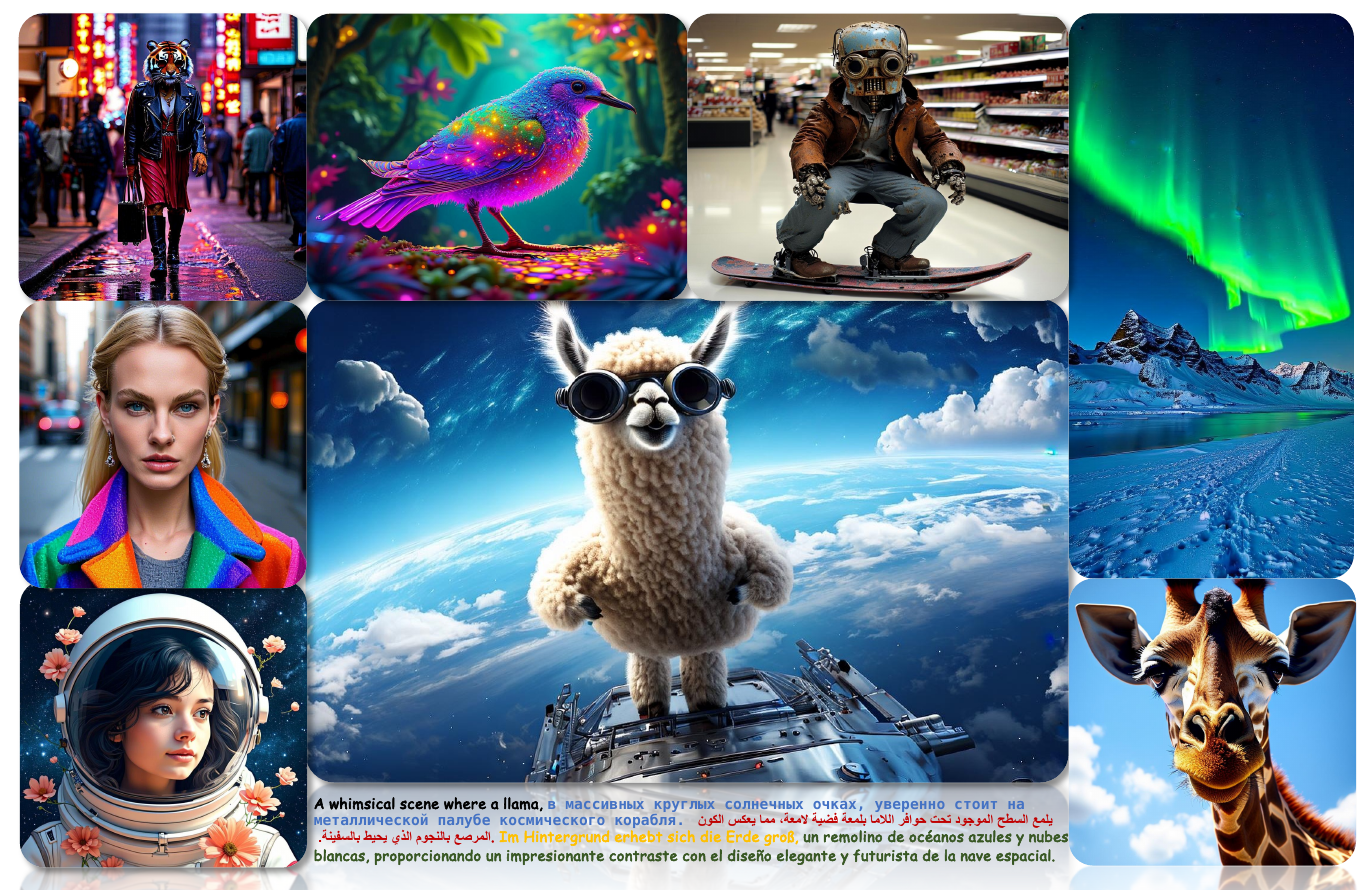}

	\caption{Generated image samples from LDGen. We present a composed prompt with each language in a different color, along with the corresponding image that exhibits high aesthetic quality and text-image alignment.}
	\label{fig:teaser}
\end{center}
}]

\let\thefootnote\relax\footnotetext{
$^\S$Corresponding author, $^\dag$ Project leader.}

\begin{abstract}
In this paper, we introduce LDGen, a novel method for integrating large language models (LLMs) into existing text-to-image diffusion models while minimizing computational demands. Traditional text encoders, such as CLIP and T5, exhibit limitations in multilingual processing, hindering image generation across diverse languages. We address these challenges by leveraging the advanced capabilities of LLMs. Our approach employs a language representation strategy that applies hierarchical caption optimization and human instruction techniques to derive precise semantic information,. Subsequently, we incorporate a lightweight adapter and a cross-modal refiner to facilitate efficient feature alignment and interaction between LLMs and image features. LDGen reduces training time and enables zero-shot multilingual image generation. Experimental results indicate that our method surpasses baseline models in both prompt adherence and image aesthetic quality, while seamlessly supporting multiple languages. Project page: \url{https://zrealli.github.io/LDGen}.

\end{abstract}

\input{sec/introduction}    
\input{sec/related}

\input{sec/method}

\input{sec/exp}

\input{sec/conclusion}


\bibliography{custom}
\input{./sec/appendix}

\end{document}

%% file: sec/introduction.tex
\section{Introduction}

Text-to-image (T2I) models aim to generate images from text descriptions.~\cite{rombach2022high,podell2023sdxl,saharia2022photorealistic,bai2024meissonic,nichol2022glide}. Thus, natural language descriptions serve as a critical bridge for conveying user intent and generating visually appealing images that accurately capture the intended semantic information. Despite the impressive performance demonstrated by advanced text-to-image models, their reliance on text encoders such as CLIP~\cite{radford2021learning} and T5~\cite{raffel2020exploring}, which are primarily tailored for English, 
constrains their multilingual capabilities due to the linguistic limitations of training datasets.

Recently, large language models~\cite{bai2023qwen,liu2024deepseek,achiam2023gpt,glm2024chatglm,dubey2024llama} have achieved notable success in the field of natural language processing. These models possess advanced language comprehension abilities, enabling them to deeply analyze prompts and provide rich, precise semantic guidance for image generation. Furthermore, many LLMs~\cite{team2024gemma,bai2023qwen,achiam2023gpt} are trained on multilingual corpora, granting them the ability to support multiple languages. These advantages have motivated researchers to explore the use of LLMs in text-to-image generation tasks. However, some prior approaches~\cite{xie2024sana,ma2024exploring,xing2024mulan,ye2024altdiffusion} have attempted to directly replace text encoders with LLMs, leading to unstable training processes and significant challenges for researchers with limited computational resources. For instance, ELLA~\cite{hu2024ella} and LLM4GEN~\cite{liu2024llm4genleveragingsemanticrepresentation} seek to align LLMs with the CLIP model but require extensive training data to adapt LLMs representations within diffusion models. These methods often treat LLMs features as mere text conditions, thereby failing to fully exploit the comprehensive language understanding capabilities of LLMs. As shown in Appendix A, directly employing LLMs for image descriptions can introduce unintended content, resulting in semantic biases and adversely affecting the output quality of diffusion models.

To effectively address these challenges and integrate large language models into existing text-to-image tasks under resource constraints, we propose LDGen. Our approach enables the efficient incorporation of LLM into current diffusion models based on T5/CLIP text encoders with minimal computational demands. As shown in Fig.~\ref{fig:pipeline}, we introduce a robust language representation strategy (LRS). By utilizing hierarchical caption optimization and human instruction strategies, LRS fully harnesses the instruction-following, in-context learning, and reasoning capabilities of LLM to accurately derive textual information, thereby enhancing semantic alignment between text and image.

Furthermore, inspired by recent advancements in alignment methods~\cite{hu2024ella,zhao2024bridging,tan2024empirical}, we employ a lightweight adapter to align LLM features with T5-XXL, substantially reducing the training time required for text-image alignment. Additionally, we introduce a cross-modal refiner to improve text comprehension and facilitate interaction between LLM and image features. After alignment, the LLM features processed through this refiner exhibit enhanced representational capability. Specifically, the cross-modal refiner integrates self-attention layers, cross-attention layers, and feed-forward neural networks.

By employing this method, LLM can be effectively integrated into existing diffusion models with minimal training. Moreover, the multilingual capabilities of LLM are preserved, enabling zero-shot multilingual image generation without the necessity for training on multilingual text-image datasets. Our experimental results demonstrate that by leveraging the intrinsic features of LLM alongside our innovative modules, LDGen surpasses the prompt comprehension performance of advanced baseline models while seamlessly supporting multiple languages. As shown in Fig.~\ref{fig:teaser}, we present several generated images. Our contributions can be summarized as follows:
\begin{itemize}[noitemsep]
    \item We present LDGen, which efficiently integrates LLM into existing text encoder-based diffusion models and supports zero-shot multilingual text-to-image generation.
    
    \item We propose a language representation strategy that leverages the capabilities of LLM through hierarchical caption optimization and human instruction strategies.
    
    \item We introduce LLM alignment and a cross-modal refiner to achieve LLM feature alignment and enhance interaction between LLM and image features, enhancing the semantic consistency of conditions.
\end{itemize}

%% file: sec/related.tex
\section{Related Work}
\begin{figure*}[t]
  \centering
  \includegraphics[width=0.92\textwidth]{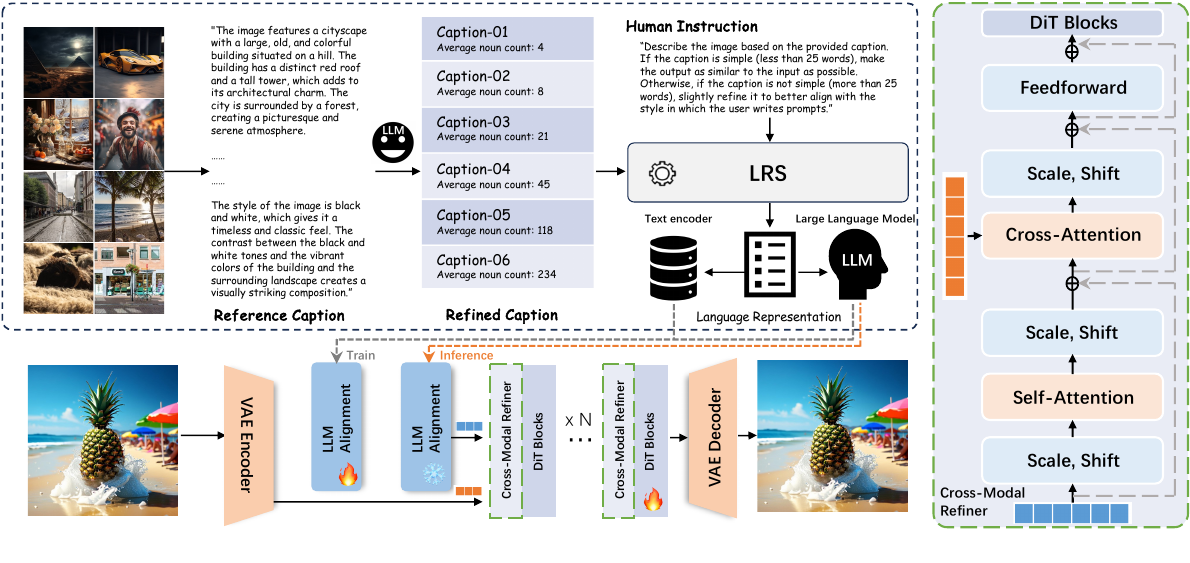}
\vspace{-1em}
  \caption{ Overview of LDGen. The dashed box shows our language representation strategy, with the bottom is our LLM alignment and cross-modal refiner training process. The detailed design of the cross-modal refiner is shown in the green box on the right.
  }
  \vspace{-1em}
  \label{fig:pipeline}
\end{figure*}

\paragraph{Text-to-Image.}

Recently, denoising diffusion probabilistic models (DDPM) \cite{ho2020denoising,nichol2021improved} have achieved breakthroughs in image synthesis and downstream applications~\cite{zhang2023adding,zhou2024magictailor,li2024photomaker,wei2023elite,li2023layerdiffusion,li2024tuning,li2024pruning,feng2024dit4edit}. By mapping the image pixels to a more compact latent space where a denoising network is trained to learn the reverse diffusion process, prominent text-guided generation models have achieved impressive results in terms of image quality and semantic fidelity. Earlier methods \cite{rombach2022high, podell2023sdxl} based on the UNet have been tremendously successful in various generative tasks. With the success of the transformer architecture in various fields, diffusion transformer-based methods~\cite{peebles2023scalable,gao2023masked} are notably developing. Techniques like FLUX \cite{flux2024} and SD3 \cite{esser2024scaling} introduced the MMBlock to further align text and images during training. PixArt-$\alpha$ \cite{chen2023pixart} explored efficient text-to-image training schemes and achieved the first Transformer-based T2I model capable of generating high-quality images at 1024 resolution. Models like Lumina-T2X \cite{gao2024lumina} and GenTron \cite{chen2024gentron} extended diffusion transformers from image generation to video generation. Playgroundv3 (PG3) \cite{liu2024playground} proposed a comprehensive VAE training, caption annotation, and evaluation strategy.

\paragraph{Large Models in T2I.}

The text encoder plays a crucial role in the text-to-image task. In the initial LDM \cite{rombach2022high}, CLIP \cite{radford2021learning} was used as the text encoder, providing the diffusion model with text comprehension capabilities. Later, Imagen \cite{saharia2022photorealistic} discovered that using a large language model with an encoder-only structure like T5 \cite{raffel2020exploring} significantly enhanced the model's text understanding. Following this, several works \cite{chen2023pixart, chen2024pixart, sun2024autoregressive, betker2023improving, esser2024scaling} utilized the T5 series of models as text encoders during pre-training. Additionally, some other works \cite{liu2024llm4genleveragingsemanticrepresentation, hu2024ella, zhao2024bridging, tan2024empirical}, attempted to adapt the T5 and LLMs~\cite{dubey2024llama} to the base models pre-trained based on CLIP. Considering the recent success of decoder-only large language models, some works have sought to apply them in image generation frameworks. PG3 \cite{liu2024playground} focused on model structure, believing that knowledge in LLMs spans all layers, thus replicating all Transformer blocks from the LLM. LiDiT \cite{ma2024exploring}, from an application perspective, designed an LLM-infused Diffuser framework to fully exploit the capabilities of LLMs. Sana \cite{xie2024sana}, focusing on efficiency, directly used the final layer of LLM features as text encoding features. Kolors \cite{kolors} adapts LLMs for use with SDXL by simply replacing the original CLIP text encoder with ChatGLM. These efforts collectively demonstrate that LLMs still hold significant research potential in the field of image generation.

%% file: sec/method.tex
\section{Method}

\subsection{Motivation}

\begin{figure}[t]
  \centering
  \includegraphics[width=0.47\textwidth]{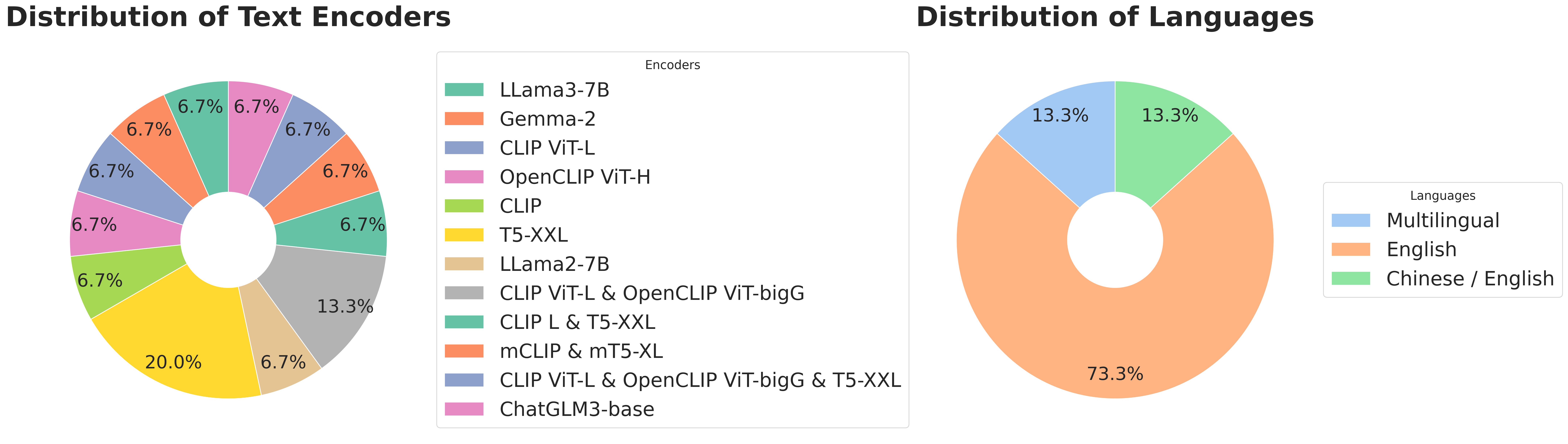}

  \caption{Distribution of text encoder and supported languages. English-based CLIP/T5 series models remain the primary text encoders.
  }
    \vspace{-0.5em}

  \label{fig:text_encoder}
  
\end{figure}

Text encoding is a pivotal component in text-to-image models, significantly influencing the quality of the generated images. As shown in Fig.~\ref{fig:text_encoder}, the CLIP~\cite{radford2021learning} and T5~\cite{raffel2020exploring} series models currently dominate the field of text encoders. However, the rapid advancement of large language models~\cite{achiam2023gpt,team2024gemma} is noteworthy. These models employ autoregressive language modeling techniques~\cite{yang2019xlnet,black2022gpt} in unsupervised learning. Through processing vast amounts of text data, they are beginning to exhibit remarkable reasoning and contextual understanding capabilities. 
They excel across a range of textual tasks. In particular, LLMs trained on multilingual corpora have demonstrated substantial promise in text-to-image generation tasks. Nonetheless, a critical challenge persists: many existing models rely on CLIP/T5 series text encoders, which are predominantly trained on English corpora and perform effectively. 
Transitioning to LLMs by replacing the existing text encoders and retraining these models from scratch would involve considerable resource expenditures. To address this issue, we employ LDGen, which seamlessly integrates LLMs into existing diffusion models based on T5/CLIP text encoders, utilizing only a small portion of the initial training resources. These new models not only outperform the originals but also enable zero-shot text-to-image generation across multiple languages.

\subsection{Language Representation Strategy}

Based on the above analysis, while large language models offer substantial advantages, they still encounter several significant challenges. As dialogue models, LLMs employing a decoder-only architecture rely on autoregressive language modeling methods. These models learn linguistic patterns through unsupervised training on large-scale text datasets by predicting the subsequent word in a sequence. However, this characteristic often makes it difficult to control the model outputs, leading to producing a lot of redundant information.

\begin{figure}[t]
  \centering
  \includegraphics[width=0.48\textwidth]{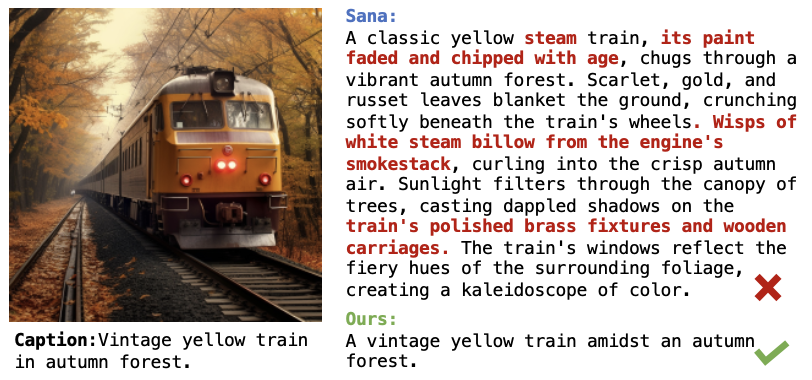}

  \caption{The red words in Sana's generated result highlight elements that do not align with the image. Providing incorrect instructions can change the original caption, potentially creating inaccurate descriptions.
  }
  \vspace{-1em}
  \label{fig:human instruction}
\end{figure}

We observe that both LiDiT~\cite{ma2024exploring} and Sana~\cite{xie2024sana} utilize human instructions to help LLMs produce more stable content. However, as shown in Fig.~\ref{fig:human instruction}, these methods can conflict with the original captions. Incorrect human instructions may cause outputs to deviate from factual accuracy and generate fabricated information, thereby disrupting text-image alignment and potentially decreasing the effectiveness of training.

To address these challenges, we employ a hierarchical captioning strategy. This approach is complemented by extensive human instruction optimization to achieve optimal language representation and enhance semantic alignment between text and images. First, similar to PG3's~\cite{liu2024playground} multi-level image description technique, we utilize the Internvl2-40B model~\cite{chen2024far,chen2024expanding} to re-caption all image data. We generate six captions of varying lengths, ranging from simple to detailed, to comprehensively capture the image content. For detailed captioning prompts, please refer to Appendix Fig.~\ref{fig:appendix_hi}, HI-05. During training, these hierarchical captions are randomly sampled and input into the LLM. As shown in Tab.~\ref{tab:hi scores}, compared to original single-caption methods, LRS enables the model to more effectively capture the hierarchical structure of language concepts while maintaining a high CLIP score.

For these complex and varied-length hierarchical captions, we further refined human instructions to ensure that the LLM's outputs maintain a high CLIP score and avoid generating non-existent information. As shown in Tab.~\ref{tab:hi scores}, the LLM surprisingly enhances the CLIP scores of the original captions, revealing that our language representation strategy effectively extracts semantic information and enhances text-to-image alignment during model training. To support multilingual text-to-image generation, we evaluated several mainstream LLMs. We selected Qwen~\cite{yang2024qwen2} as our preferred model because it is one of the few trained on multilingual corpora and exhibits exceptional performance in text-related tasks.

\definecolor{pz}{RGB}{230,230,230}
\begin{table*}[h]
\centering

\caption{Human Instruction Comparison. Each entry has the CLIP-Score~\cite{hessel2021clipscore} on the left and the LongCLIP-Score~\cite{zhang2024long} on the right, with the average word number is in \textcolor{gray}{gray brackets (.)}. \textbf{Original} refers to the initial caption. "HI" indicates outputs from various Human Instruction strategies. The highest scores are highlighted in \textbf{bold}, while the second-highest scores are \underline{underlined}. Scores that surpass the original captions are marked with a \colorbox{pz}{gray background}.}
\renewcommand{\arraystretch}{1.1}
\resizebox{\textwidth}{!}{
\begin{tabular}{lcccccccc}
\toprule[1.2pt]
 & \textbf{Original} & \textbf{Ours} & \textbf{No-HI} & \textbf{HI-01} & \textbf{HI-02} & \textbf{HI-03} & \textbf{HI-04} & \textbf{HI-05} \\
\hline
Caption-1 & 27.65/29.53 &  \colorbox{pz}{\textbf{27.66}}/\colorbox{pz}{\underline{29.74}} & 22.21/27.44 & 22.79/27.95 & \underline{23.33}/\textbf{30.04} & 22.61/26.59 & 23.06/29.61 & 22.40/27.97 \\
& \textcolor{gray}{(4.48)} & \textcolor{gray}{(7.63)} & \textcolor{gray}{(204.29)} & \textcolor{gray}{(335.19)} & \textcolor{gray}{(87.52)} & \textcolor{gray}{(171.14)} & \textcolor{gray}{(254.19)} & \textcolor{gray}{(224.43)} \\
 \hline
 
Caption-2 & 29.65/31.49 & \colorbox{pz}{\textbf{29.66}}/\colorbox{pz}{\underline{31.63}} & 22.89/28.35 & 23.76/30.31 & \underline{24.47}/\textbf{31.78} & 23.32/28.89 & 24.07/31.25 & 23.41/29.93 \\
& \textcolor{gray}{(8.99)} & \textcolor{gray}{(11.67)} & \textcolor{gray}{(173.66)} & \textcolor{gray}{(307.68)} & \textcolor{gray}{(70.00)} & \textcolor{gray}{(172.76)} & \textcolor{gray}{(245.64)} & \textcolor{gray}{(214.01)} \\
\hline

Caption-3 & 30.20/33.24 & \textbf{29.50}/\textbf{32.92} & 23.75/29.52 & 24.40/31.05 & \underline{25.29}/\underline{32.77} & 24.31/30.31 & 24.72/32.37 & 24.33/31.03 \\
& \textcolor{gray}{(21.71)} & \textcolor{gray}{(25.25)} & \textcolor{gray}{(183.91)} & \textcolor{gray}{(306.33)} & \textcolor{gray}{(80.78)} & \textcolor{gray}{(194.68)} & \textcolor{gray}{(226.49)} & \textcolor{gray}{(192.64)} \\
 \hline
 
Caption-4 & 27.53/34.64 & \textbf{27.13}/\textbf{33.76} & 24.56/30.03 & 24.67/31.49 & \underline{25.33}/\underline{33.35} & 24.63/30.42 & 25.01/33.25 & 24.89/32.19 \\
& \textcolor{gray}{(45.16)} & \textcolor{gray}{(46.96)} & \textcolor{gray}{(249.35)} & \textcolor{gray}{(329.49)} & \textcolor{gray}{(116.45)} & \textcolor{gray}{(253.19)} & \textcolor{gray}{(205.26)} & \textcolor{gray}{(167.07)} \\
 \hline
 
Caption-5 & 25.39/34.43 & \colorbox{pz}{\textbf{25.40}}/\colorbox{pz}{\textbf{33.74}} & 23.87/30.33 & 24.86/31.78 & \underline{25.37}/33.37 & 24.38/29.65 & 25.22/\underline{33.52} & 25.30/33.20 \\
& \textcolor{gray}{(118.06)} & \textcolor{gray}{(106.18)} & \textcolor{gray}{(304.45)} & \textcolor{gray}{(350.39)} & \textcolor{gray}{(183.10)} & \textcolor{gray}{(334.34)} & \textcolor{gray}{(205.27)} & \textcolor{gray}{(177.34)} \\
 \hline
 
Caption-6 & 25.42/34.65 & \colorbox{pz}{\underline{25.48}}/\textbf{33.96} & 23.72/31.09 & 25.05/32.18 & 25.36/33.60 & 24.26/30.45 & 25.38/33.89 & \textbf{25.59}/\underline{33.91} \\
& \textcolor{gray}{(118.06)} & \textcolor{gray}{(106.18)} & \textcolor{gray}{(304.45)} & \textcolor{gray}{(350.39)} & \textcolor{gray}{(183.10)} & \textcolor{gray}{(334.34)} & \textcolor{gray}{(205.27)} & \textcolor{gray}{(177.34)} \\
\toprule[1.2pt]
\end{tabular}
}
\label{tab:hi scores}
\end{table*}

\begin{figure*}[t]
	\centering
	\includegraphics[width=0.98\linewidth]{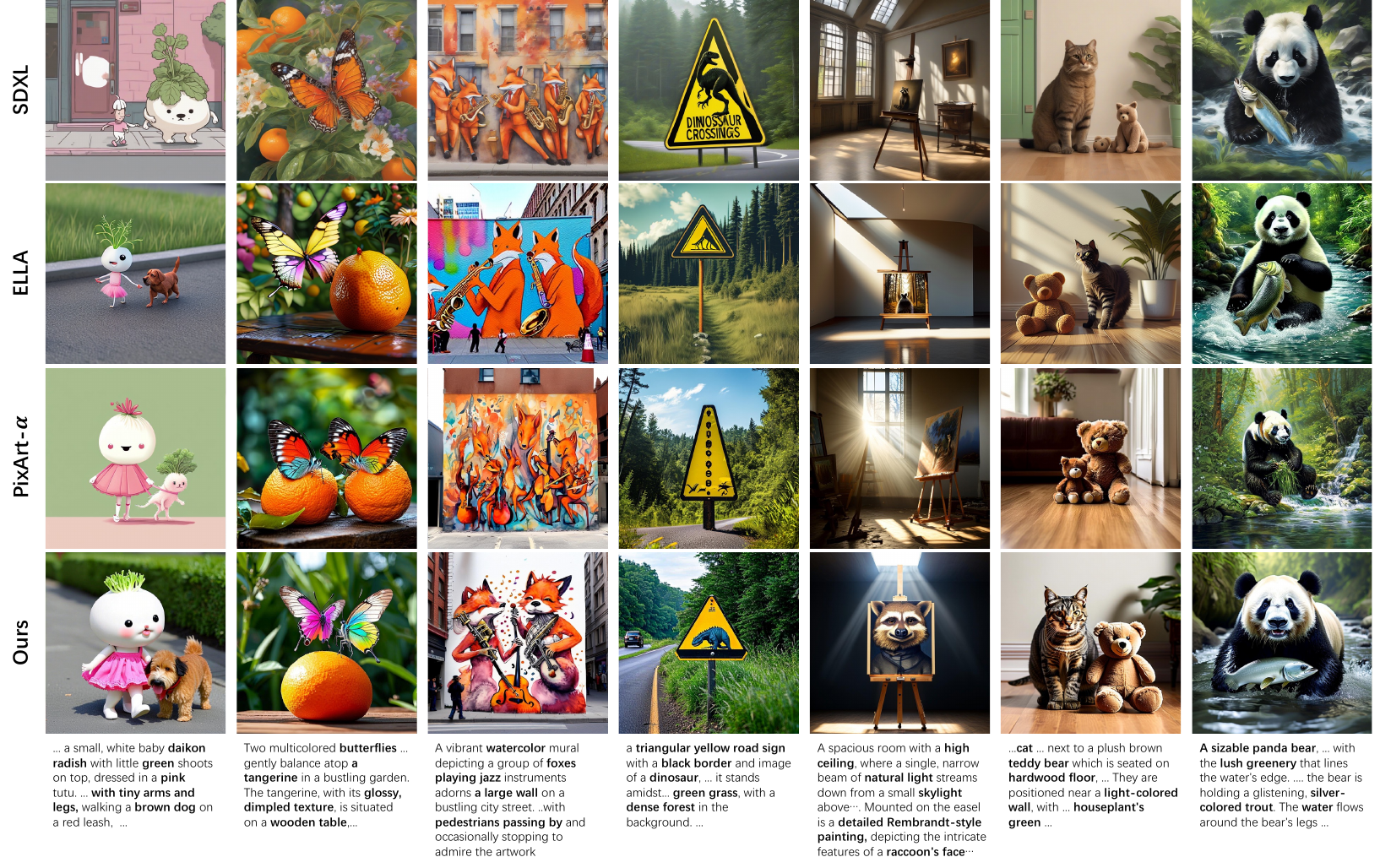}

	\caption{Comparison of our method with recent enhancement generative models ELLA~\cite{hu2024ella}, baseline Models SDXL~\cite{podell2023sdxl} and PixArt-$\alpha$~\cite{chen2023pixart}. Our method achieves the best results in terms of instruction adherence and visual appeal.}
	\label{fig:comparasions}
\end{figure*}

\subsection{LLM Alignment}

For pre-trained diffusion models~\cite{chen2023pixart,podell2023sdxl}, aligning the original text encoder with LLMs features using linear layers is challenging. This is primarily due to the significant differences in the output feature spaces of T5/CLIP encoders and LLMs. As a result, directly modifying and training the existing model structure can lead to instability. To address this, we employ a two-step approach: first, we align the feature spaces, then fine-tune the model weights to adapt to the new feature space. This method significantly reduces training time.

Specifically, we first multiply the LLM output by a small coefficient to match the numerical range of T5. This effectively speeds up the feature alignment training. Next, similar to previous methods~\cite{tan2024empirical}, we design a three-layer encoder-decoder Transformer adapter to align the feature spaces of the T5 encoder and LLM output. During the adapter training, we utilize the following alignment loss functions: $\lambda_{1}* \mathcal{L}_{\text{cos}} + \lambda_{2}*\mathcal{L}_{\text{MSE}}$. The cosine similarity loss aligns the feature space directions, and mean squared error (MSE) loss can further enhance alignment accuracy in terms of numerical range.

By optimizing the alignment loss, we achieve a rough alignment between LLM and T5 output feature spaces. This allows us to quickly integrate the LLM into the pre-trained diffusion model, enhancing its overall performance and adaptability.

\subsection{Cross-Modal Refiner}

To improve text comprehension and facilitate interaction between LLM features and image features, we introduce a lightweight module called the cross-modal refiner. This module employs a sequence of components to optimize and refine LLM feature representations, enabling efficient integration of text and image features. As shown in Fig.~\ref{fig:pipeline}, it includes elements such as self-attention mechanisms, cross-attention mechanisms, feedforward neural networks, residual connections, normalization layers, and learnable scaling factors.

To enhance the interaction between image and text features, the cross-attention layer serves as a pivotal component of modal interaction. This layer utilizes LLM features as queries, with latent image features acting as keys and values, to facilitate deep interaction between text and image elements. This design enables the refinement and adjustment of text features based on relevant image information, thereby enhancing the model's understanding of cross-modal content. Learnable scaling factors allow the model to gradually balance between original and optimized features during training, ensuring a seamless transition from pre-trained weights to new LLM input features. This mechanism effectively integrates the original LLM's robust semantic understanding into the pre-trained models, boosting overall performance.

The cross-modal refiner module preserves the original LLM features and effectively integrates image-related information to produce richer, semantically aligned conditional representations. This approach allows us to efficiently integrate the LLM into existing diffusion models within relatively short training times, providing highly semantically aligned conditional information for text-to-image generation tasks, significantly enhancing the quality and relevance of generated results.

%% file: sec/exp.tex
\section{Experiments}

\begin{table}[t]
\scriptsize
\centering
\renewcommand{\arraystretch}{1.15}
\setlength{\tabcolsep}{4pt}

\caption{Quantitative comparison results on DPG-Bench. Note that we support multiple languages.}
\label{tab:dpg}

\resizebox{\linewidth}{!}{%
\begin{tabular}{lcccccc}
\toprule[0.8pt]

\multirow{1}{*}{\textbf{Method}} &\multicolumn{1}{c}{\textbf{Param}}& \multicolumn{1}{c}{\textbf{Multi-Ling}} & \multicolumn{1}{c}{\textbf{DPG-Bench}} \\ \hline

SD1.5 \cite{rombach2022high} &${0.86B}$& $\times$&${61.18}$  \\ 
SDv2.1 \cite{rombach2022high} &${0.89B}$& $\times$&${68.09}$  \\ 
LlamaGen \cite{sun2024autoregressive} &${0.78B}$& $\times$&${65.16}$  \\ 

HART \cite{tang2024hart} &${0.73B}$& $\times$&${80.89}$  \\ 

Sana \cite{xie2024sana} &${0.60B}$& $\checkmark$&$83.6$  \\ 
ELLA \cite{hu2024ella} &${0.93B}$& $\times$&${80.79}$  \\ 
LLM4GEN\cite{liu2024llm4genleveragingsemanticrepresentation} &${0.86B}$& $\times$&${67.34}$  \\ 
Pixart-$\alpha$ \cite{chen2023pixart} &${0.61B}$& $\times$&${71.11}$  \\ 
Ours  &${0.63B}$&$\checkmark$& ${80.57}$  \\  
\midrule
SD3-Medium \cite{esser2024scaling} &${2.0B}$&$\times$& ${84.08}$  \\ 
SDXL \cite{podell2023sdxl} &${2.6B}$&$\times$& ${74.65}$  \\ 
Janus \cite{wu2024janus} &${1.3B}$& $\times$&${79.68}$  \\ 
Janus-Pro \cite{chen2025janus} &${7B}$& $\times$&${84.19}$  \\ 
Emu-3 \cite{wang2024emu3} &${8.0B}$& $\times$&${80.60}$  \\ 
 DALL-E 3 \cite{betker2023improving} &${-}$&$\times$& ${83.50}$  \\ 

FLUX-Dev \cite{flux2024} &${12.0B}$& $\times$&${84.0}$  \\ 

\toprule[0.8pt]

\end{tabular}%
}

\end{table}

\begin{figure*}[t]
	\centering
	\includegraphics[width=0.95\linewidth]{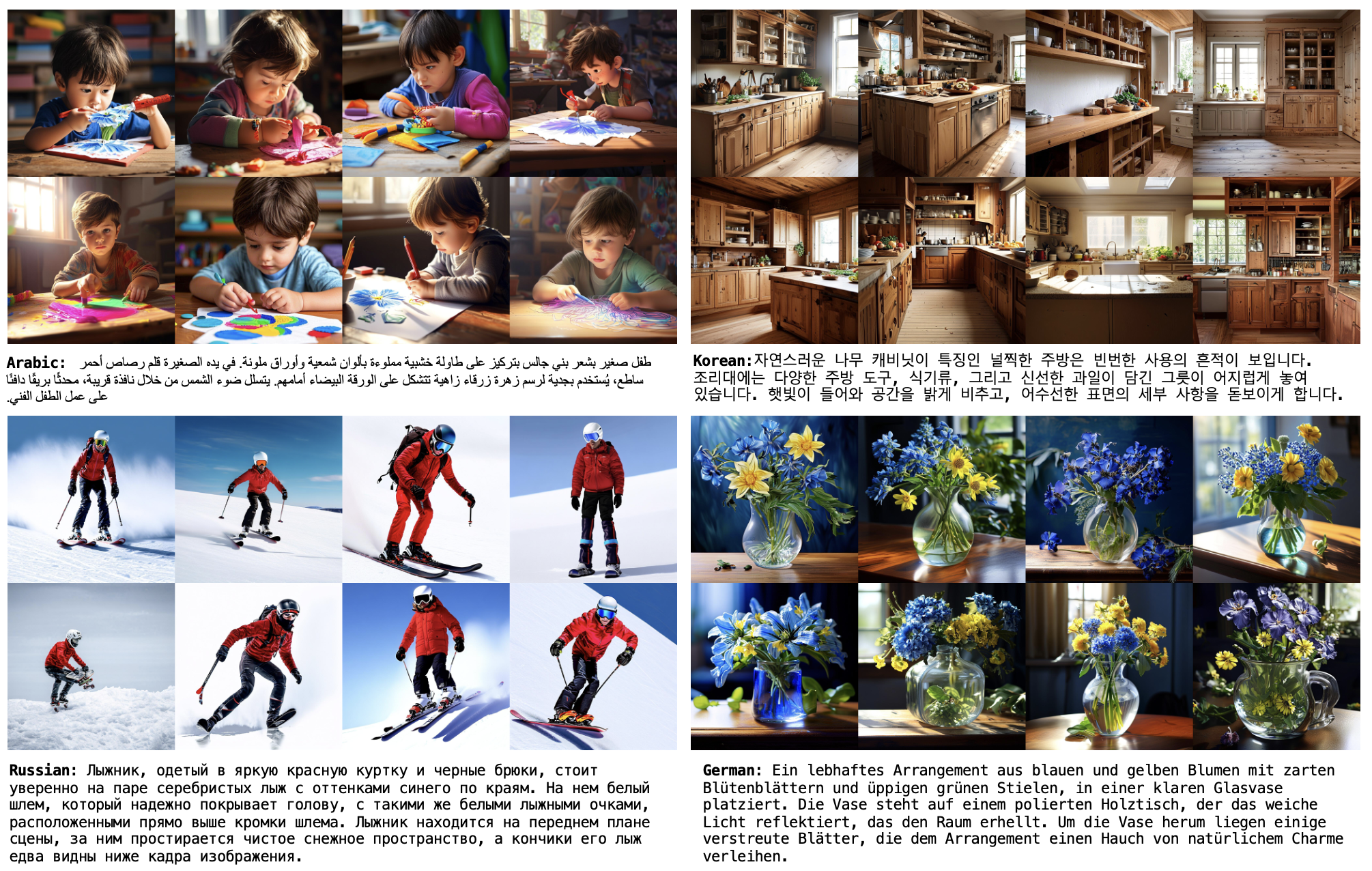}

	\caption{Multilingual qualitative visualization results. For each panel's eight images, we generate them using eight different languages but only display the prompt in one of the languages used. Note that LDGen uses only English prompts during training but achieves zero-shot multilingual generation due to the capabilities of the LLM.}
	\label{fig:multi-ling}
\end{figure*}

\begin{table*}[t]
\centering
    \caption{We compare our method with baseline methods and fine-tuned baseline methods on DPG-Bench and Geneval, demonstrating the effectiveness of our approach.}
\renewcommand{\arraystretch}{1}
\setlength{\tabcolsep}{5pt}
    \resizebox{\linewidth}{!}{
    
    \begin{tabular}{lccccccccccccc}
    \toprule[1pt]
    \multirow{2}{*}{Method}&\multicolumn{1}{c}{Param} & \multicolumn{5}{c}{DPG-Bench \cite{hu2024ella}}  & \multicolumn{5}{c}{Geneval\cite{ghosh2023geneval}}  \\
    \cmidrule(lr){3-7} \cmidrule(lr){8-12} 
    && Global & Entity & Attri. & Other &Overall  & Single Obj. & Two Obj. & Counting & Color Attri. & Overall\\
    \midrule

    Pixart-$\alpha$~\cite{chen2023pixart}& 0.61B & 74.97 & 79.32 & 78.60 & 76.69 & 71.11 & \textbf{0.98} & 0.50 & \textbf{0.44} & 0.07 & 0.48  \\
    Pixart-$\alpha$(fine-tuned) & 0.61B& 83.18 & 84.06 & 84.07 & 83.61 & 75.05& 0.95 & 0.37 & 0.37 & 0.43 & 0.46  \\
    Ours & 0.63B& \textbf{85.88} & \textbf{87.83} & \textbf{85.21} & \textbf{87.85} & \textbf{80.57} & 0.88 & \textbf{0.55} &  0.35& 0.42 & \textbf{0.51} \\
    \bottomrule
    \end{tabular}
    }
  
\label{Tab: com_pixart}
\end{table*}

\paragraph{Model Details.}

Our method is based on the work of PixArt-$\alpha$~\cite{chen2023pixart}, which is a classic diffusion transformer text-to-image model. It uses the T5-XXL text encoder~\cite{raffel2020exploring} and has demonstrated excellent performance. We use Qwen2.5-7B-Instruct~\cite{yang2024qwen2} as the LLM and adopt the output features from the last layer, which has a dimension of 3584. The VAE remains consistent with PixArt-$\alpha$~\cite{chen2023pixart}. For the LLM feature alignment module, we employ a 3-layer encoder-decoder transformer structure, which includes linear layers to align the LLM dimension of 3584 with the T5 dimension of 4096. The cross-modal refiner uses only one block.
\paragraph{Training Details.}

To reduce computational resources, we've structured our training process into several key stages. First, we train the LLM feature alignment module using approximately 80 million text entries from internal image descriptions, with about 20\% of this data being multilingual. Given that T5-XXL~\cite{raffel2020exploring} doesn't support multiple languages, we align the multilingual features from the LLM output with the English output features of T5-XXL~\cite{raffel2020exploring}. This initial phase consumes around 80 A100 GPU days. Next, drawing inspiration from PixArt-$\alpha$'s training methodology, we adapt our model to 512 resolution and fine-tune it using 24 million text-image pairs. To minimize dataset-specific biases in training, we maintain a data scale similar to PixArt-$\alpha$'s~\cite{chen2023pixart} original approach and incorporate various datasets with overlapping ranges, such as JourneyDB~\cite{sun2023journeydb}. In the final stage, we continue training at a 1024 resolution, utilizing 14 million aesthetic data entries. The entire training process requires approximately 120 A100 GPU days. The count of GPU days excludes the time for T5, Qwen, and VAE feature extraction. LDGen takes only approximately 26\% of the GPU days compared to PixArt-$\alpha$.

\paragraph{Evaluation Metrics.}

We evaluate our approach using two publicly available benchmarks: Geneval~\cite{ghosh2023geneval} and DPG-Bench~\cite{hu2024ella}. Geneval is a challenging text-to-image generation benchmark designed to showcase a model's comprehensive generative capabilities through detailed instance-level analysis. DPG-Bench, comprises 1,065 semantically dense long prompts, aimed at evaluating model performance in complex semantic alignment. 

These two datasets provide a comprehensive assessment of generative models from different perspectives.

\subsection{Performance Comparison and Analysis}

We focus on evaluating the performance of our method compared to the baseline model, PixArt-$\alpha$~\cite{chen2023pixart}. As shown in Tab.~\ref{tab:dpg} and Tab.~\ref{Tab: com_pixart}, we utilize two evaluation benchmarks, DPG-Bench~\cite{hu2024ella} and Geneval~\cite{ghosh2023geneval}, to thoroughly assess image-text consistency.

Furthermore, we compare our results with advanced models such as the Stable Diffusion series and enhancement methods like ELLA~\cite{hu2024ella} and LLM4GEN~\cite{liu2024llm4genleveragingsemanticrepresentation}. Our model not only surpasses these baseline models but also achieves approximately a 13\% performance improvement on DPG-Bench compared to PixArt-$\alpha$, approaching the metrics of some larger-scale models. For the Geneval results, we notice that while single-object scores might decrease due to the LLM's data alignment scale being significantly smaller than the hundreds of millions of samples used for text encoder training, we see significant improvements in multiple aspects, such as color attributes, with the LLM's use.

Although we have made progress, there remains a gap when compared to state-of-the-art models such as HART~\cite{tang2024hart} and Sana~\cite{xie2024sana}, which are trained from scratch with extensive resources and incorporate cutting-edge techniques. Nevertheless, our method achieves significant performance gains on the base model with relatively minimal overhead. Tab.~\ref{tab:dpg} presents our evaluation scores across different languages. Even without using multilingual image-text pairs during training, our model achieves a score of 61.3 in some common languages, nearly matching the 61.2 of certain English-trained image generation models (like SD1.5~\cite{rombach2022high}). As shown in Tab.~\ref{tab:multi-lang}, we conduct a multilingual generation comparison with Sana and additionally support languages that are not supported by Sana.

As shown in Fig.~\ref{fig:comparasions}, we present visual comparisons with other enhancement methods like ELLA~\cite{hu2024ella} and LLM4GEN~\cite{liu2024llm4genleveragingsemanticrepresentation}, as well as the baseline PixArt-$\alpha$. Our method exhibits significant improvements in both aesthetics and text alignment, attributed to the integration of an LLM model with robust comprehension capabilities. Even without employing multilingual image-text data during fine-tuning, our model can generate aesthetically pleasing, instruction-following images in multiple languages.

As shown in Fig.~\ref{fig:multi-ling}, we present generation results in eight languages, displayed from top left to bottom right: German, Spanish, Portuguese, Russian, Italian, Korean, English, and Arabic. Although the model may not generate high-fidelity details across different languages, it is still capable of creating many common scenes and objects.

\begin{table}[t]
\scriptsize
\centering
\renewcommand{\arraystretch}{1.0}
\setlength{\tabcolsep}{2.8pt}
\caption{Quantitative comparisons of multilingual generation results.
We additionally support some languages that are not supported by Sana. }
\label{tab:multi-lang}

\resizebox{\linewidth}{!}{%
\begin{tabular}{lcccccc}
\toprule[0.8pt]

\multirow{1}{*}{\textbf{Language}} &\multicolumn{1}{c}{\textbf{Overall$\uparrow$}} & \multicolumn{1}{c}{Glob.} & \multicolumn{1}{c}{Enti. } & \multicolumn{1}{c}{Attr.} & \multicolumn{1}{c}{Rela. } & \multicolumn{1}{c}{Other} \\ \hline

Korean (Sana) &$10.6$& $20.3$ & $21.3$ & $20.1$ &  $20.5$ &$23.7$  \\ 
Korean  (Ours)&$\textbf{50.5}$& $\textbf{73.8}$ & $\textbf{63.6}$ & $\textbf{68.1}$ &  $\textbf{70.4}$ &$\textbf{68.6}$  \\ 
\midrule

Arabic (Sana) &$12.5$& $22.1$ & $26.1$ & $23.8$ &  $25.4$ &$31.2$  \\ 
Arabic (Ours) &$\textbf{50.0}$& $\textbf{64.4}$ & $\textbf{66.3}$ & $\textbf{66.4}$ &  $\textbf{72.9}$ &$\textbf{66.5}$  \\ 

\midrule

Russian (Sana) &$42.2$& $57.5$ & $57.2$ & $56.6$ &  $59.7$ &$62.2$  \\ 
Russian (Ours)&$\textbf{55.9}$& $\textbf{76.1}$ & $\textbf{70.8}$ & $\textbf{71.4}$ &  $\textbf{73.5}$ &$\textbf{70.2}$  \\ 
\midrule

Spanish (Sana) &$\textbf{67.4}$& $\textbf{78.9}$ & $\textbf{78.1}$ & $\textbf{79.6}$ &  $79.8$ &$75.3$  \\ 
Spanish (Ours) &$61.3$& $74.1$ & $72.0$ & $76.7$ &  $\textbf{80.3}$ &$\textbf{77.9}$  \\

\toprule[0.8pt]

\end{tabular}%
}

\end{table}

\subsection{Ablation Study}

In this section, we validate our language representation strategy, LLM alignment module, and cross-modal refiner. First, we conduct a detailed ablation analysis of our Human Instruction (HI) design, with specific details provided in the appendix. Some captions' length exceed CLIP's evaluation capacity, but with LongCLIP~\cite{zhang2024long} supporting up to 248 tokens, we use the LongCLIP score as an additional metric. We randomly select 5,000 samples from the training dataset for calculating their CLIPScore~\cite{hessel2021clipscore} and LongCLIP-Score. As shown in Tab.~\ref{tab:hi scores}, our HI strategy significantly enhances the CLIP scores of the original captions, demonstrating that our language representation strategy accurately extracts text embeddings and effectively improves text-image alignment during model training.

Although our training data size is similar to PixArt-$\alpha$, to eliminate the potential benefits of extra data, we fine-tune the original PixArt-$\alpha$ weights using the T5-XXL~\cite{raffel2020exploring} with the same training data. As shown in Tab.~\ref{Tab: com_pixart}, our method remains superior to this fine-tuned model, validating the effectiveness of our LLM alignment module and cross-modal refiner.

\section{Conclusion}

This paper presents LDGen, which integrates LLMs with diffusion models to enhance text-to-image generation. By using the language representation strategy, LLM alignment module, and cross-modal refiner, we improve semantic alignment between text and images, reduce training demands, and enable zero-shot multilingual generation. Experiments indicate the superiority of LDGen and provide new insights into LLM-T2I tasks.

\section{Limitations}

Our work integrates LLM into diffusion models with text encoders, enhancing text-image alignment and enabling excellent zero-shot multilingual image generation using limited resources. However, our LLM alignment training data is smaller compared to classic text encoders, potentially affecting the understanding of complex prompts and alignment for certain concepts. Additionally, uneven multilingual corpora distribution leads to varied performance across languages. We plan to expand training data in the future to address these issues.

%% file: sec/appendix.tex
\clearpage

\appendix
\section{Appendix}
\begin{figure*}[b]
  \centering
  \includegraphics[width=0.95\textwidth]{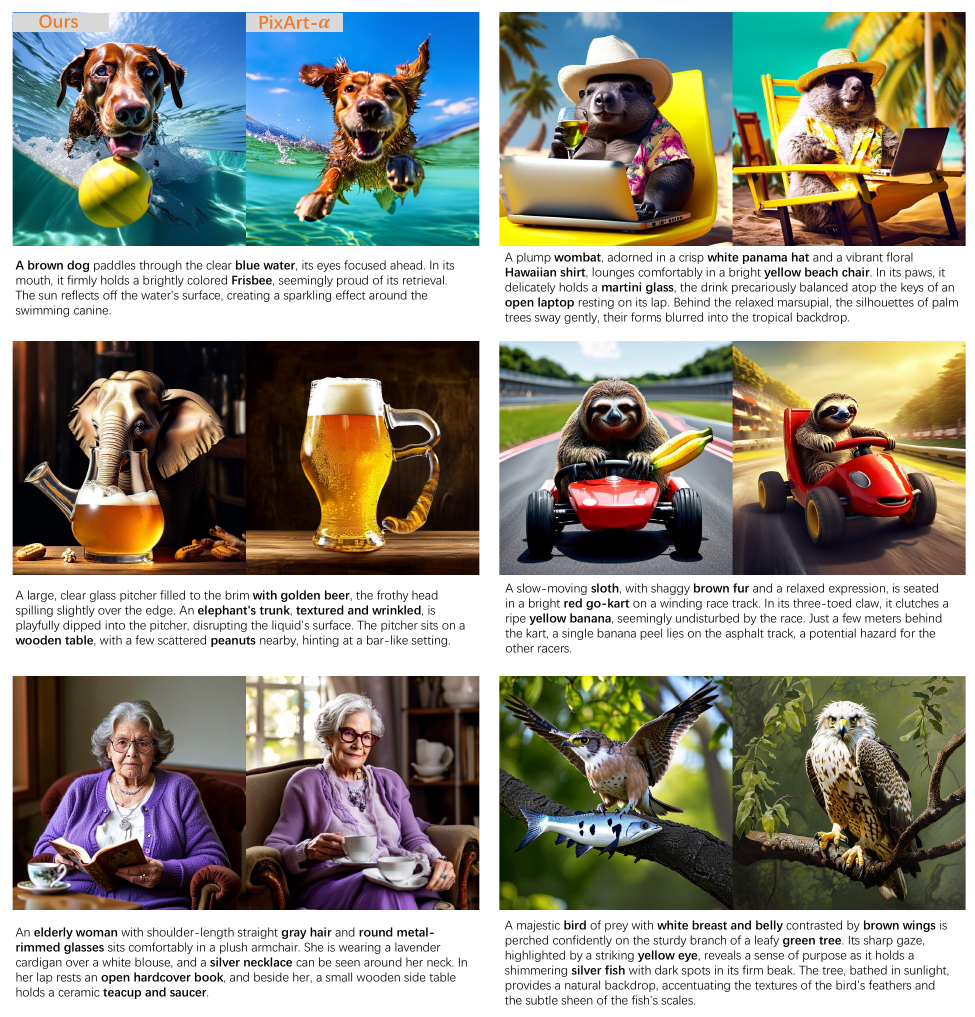}

  \caption{ More comparisons with Pixart-$\alpha$. Our method achieves better results in terms of prompt adherence and visua appeal.}
  \label{fig:appendix_com_pixart}
\end{figure*}

In the appendix, we provide a more comprehensive analysis of the main text, enriched with additional details to enhance understanding.

In Fig.~\ref{fig:appendix_com_pixart}, we provide a detailed comparison between our method and the baseline method, Pixart-$\alpha$~\cite{chen2023pixart}, which demonstrates weaker text comprehension capabilities. Our approach shows significant improvements in terms of aesthetic quality and prompt adherence.

\begin{figure*}[t]
  \centering
  \includegraphics[width=0.85\textwidth]{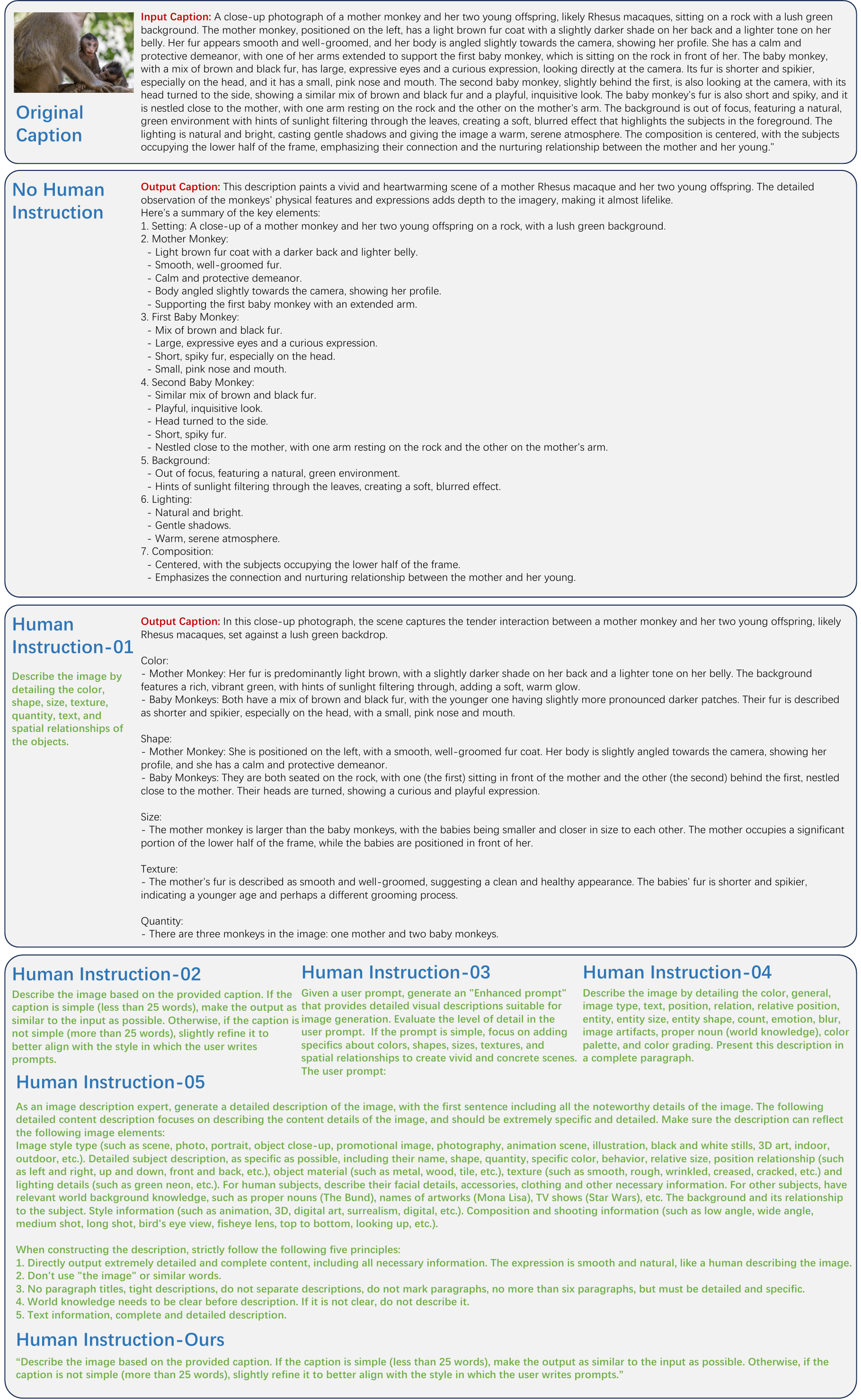}
  \caption{ We provide detailed comparisons using human instructions ranging from simple to complex, comprehensively evaluating the effectiveness of our method.
  }
  \label{fig:appendix_hi}
\end{figure*}

In Fig~\ref{fig:appendix_hi}, we perform an extensive comparison utilizing the Human Instruction with the Qwen2.5-7B-Instruct~\cite{yang2024qwen2} of a large language model (LLM), consistent with the version applied in the main text. Our Human Instruction method ensures that the LLM's outputs not only sustain a high CLIP score but also avoid generating non-existent information. Furthermore, this method enhances the accuracy of text embeddings, leading to more reliable outcomes.

\begin{figure*}[t]
  \centering
  \includegraphics[width=0.99\textwidth]{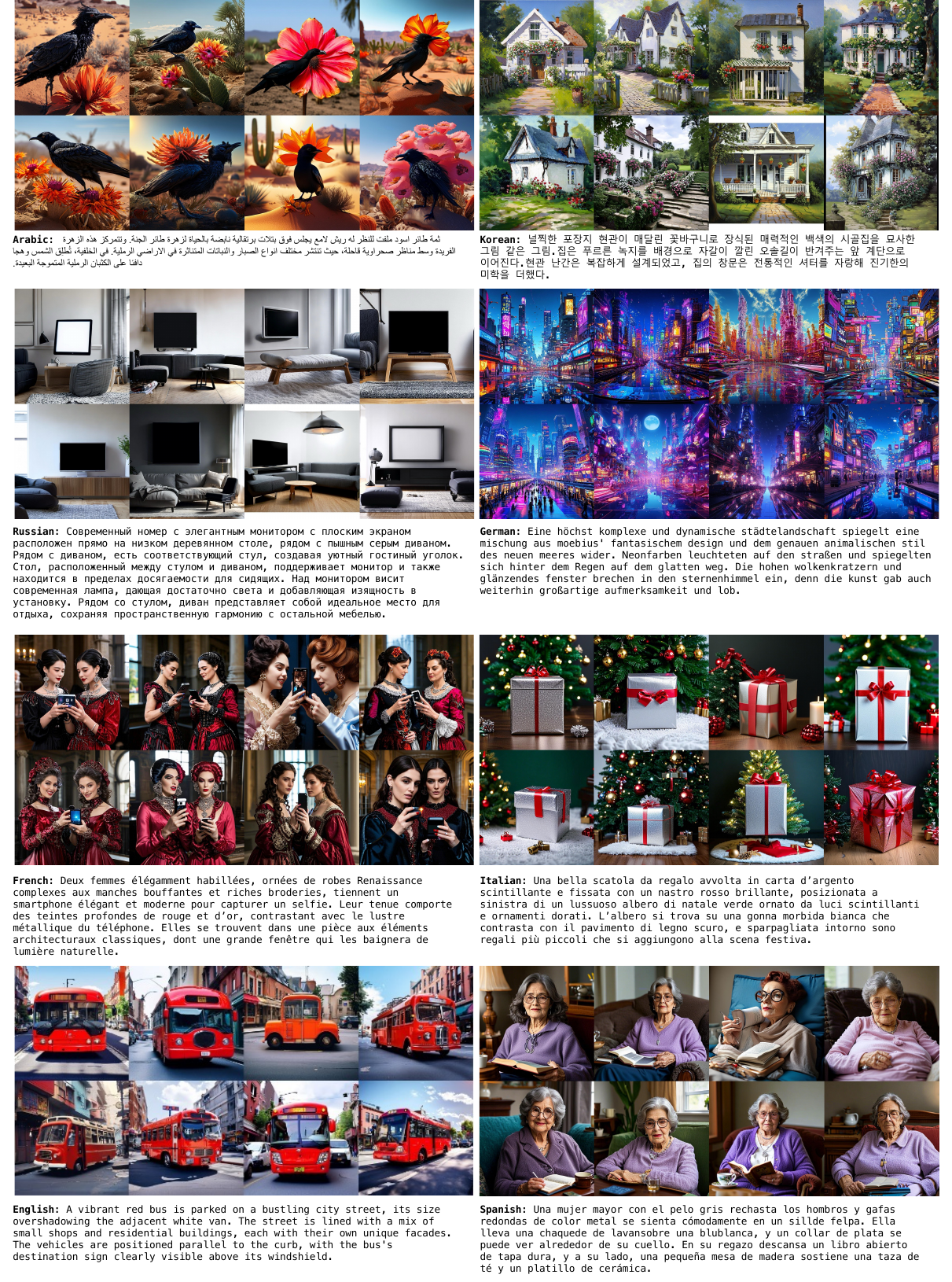}

  \caption{ More multilingual qualitative visualization results. For each panel's eight images, we generate them using eight different languages but only display the prompt in one of the languages used. 
  }
  \label{fig:appendix_multi}
\end{figure*}

Fig.~\ref{fig:appendix_multi} displays more multilingual generation results. Although some images show slight deficiencies in adhering to the prompts, they still produce outstanding results for many common scene.

\begin{figure*}[t]
  \centering
  \includegraphics[width=0.99\textwidth]{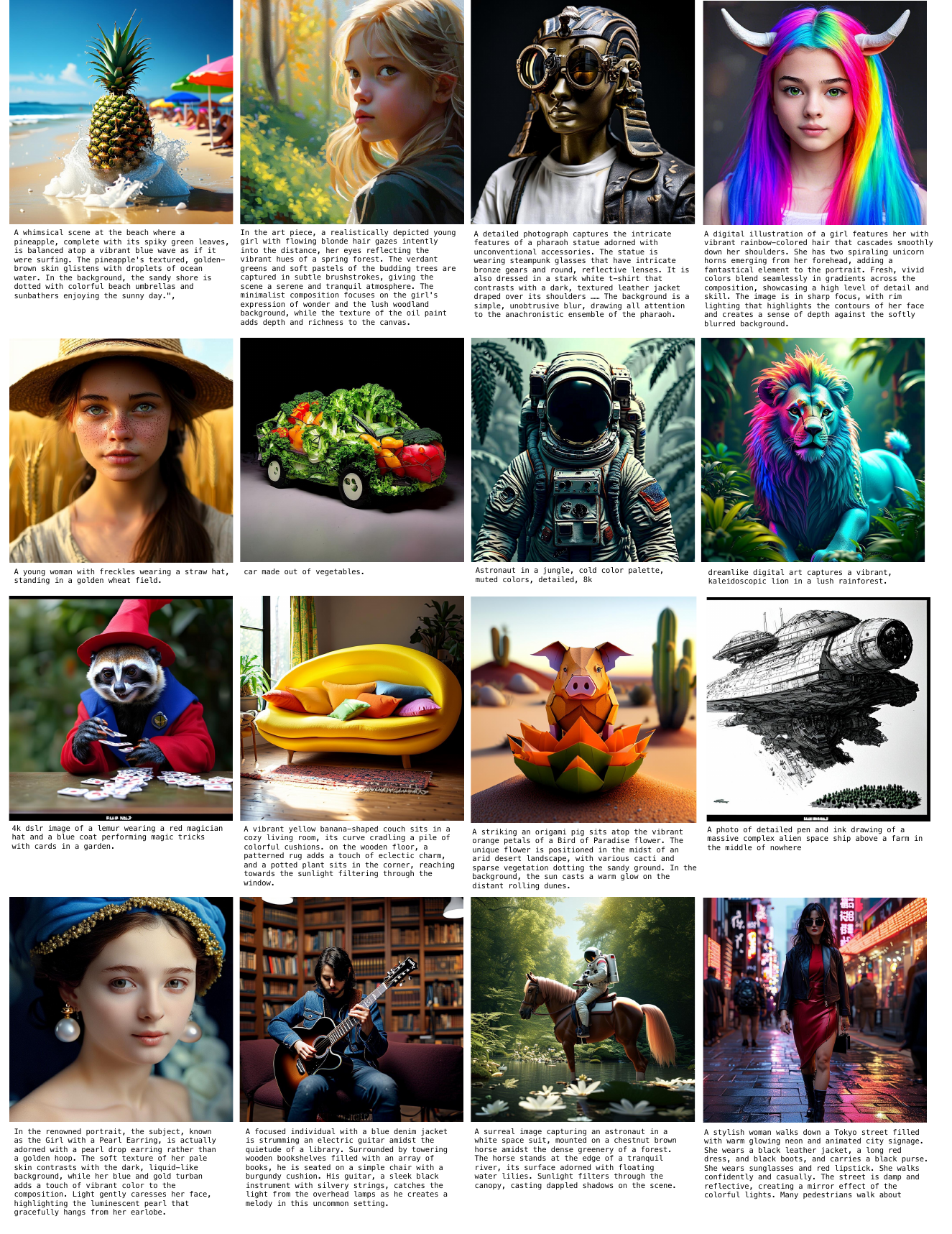}

  \caption{ More samples generated from LDGen.
  }
  \label{fig:appendix_more_vis}
\end{figure*}

Fig.~\ref{fig:appendix_more_vis} showcases images produced from multiple perspectives, including color, theme, style, etc. These varied perspectives effectively illustrate the effectiveness and adaptability of LDGen,

